\documentclass[letterpaper, 10 pt, conference]{ieeeconf}  

\IEEEoverridecommandlockouts                              

\title{\LARGE \bf
Manipulation Planning Among Movable Obstacles Using Physics-Based Adaptive Motion Primitives
}

\author{Dhruv Mauria Saxena*, Muhammad Suhail Saleem*, and Maxim Likhachev
\thanks{*Dhruv Mauria Saxena and Muhammad Suhail Saleem contributed equally to this work.}
\thanks{The authors are with the Robotics Institute, Carnegie Mellon University, Pittsburgh, PA 15213, USA. {\small e-mail: \tt \{dsaxena, msaleem2, mlikhach\}@andrew.cmu.edu}}%
}

\usepackage{xspace}
\usepackage{hyperref}

\newcommand\moduleName[1]{\textsf{#1}\xspace}
\newcommand{\MAMO}{\moduleName{MAMO}}
\newcommand{\NAMO}{\moduleName{NAMO}}
\newcommand{\AMP}{\moduleName{AMP}}

\usepackage[table,xcdraw,dvipsnames,x11names]{xcolor}
\usepackage{color}
\definecolor{burgundy}{rgb}{0.5, 0.0, 0.13}
\def\D#1{\textcolor{black}{#1}}

\newcommand{\PP}{\mathcal{P}}
\newcommand{\X}{\mathcal{X}}
\newcommand{\A}{\mathcal{A}}
\newcommand{\T}{\mathcal{T}}

\newcommand{\R}{\mathcal{R}}
\newcommand{\Obs}{\mathcal{O}}

\usepackage{graphicx}
\usepackage{amsmath}
\usepackage{amssymb}

\usepackage{amsthm}

\theoremstyle{definition}
\newtheorem{definition}{Definition}[]

\theoremstyle{assumption}
\newtheorem{assumption}{Assumption}[]

\DeclareMathOperator*{\argmin}{arg\,min}

\usepackage{siunitx}
\usepackage[normalem]{ulem}

\usepackage{algorithm}
\usepackage[noend]{algpseudocode}
\let\oldReturn\Return
\renewcommand{\Return}{\State\oldReturn}

\usepackage{multirow}
\usepackage{booktabs}

\definecolor{ObsRed}{RGB}{211,95,95}
\definecolor{ObsGreen}{RGB}{113,200,55}
\definecolor{EllipseBlue}{RGB}{55,171,200}
\definecolor{CircleOrange}{RGB}{255,102,0}
\definecolor{CirclePurple}{RGB}{128,0,128}

\usepackage{geometry}
\begin{document}

\maketitle
\thispagestyle{empty}
\pagestyle{empty}

\begin{abstract}

Robot manipulation in cluttered
scenes often requires contact-rich interactions with objects. It can be more economical to interact
via non-prehensile actions, for example, push through other objects to get to the desired grasp pose, instead of deliberate prehensile rearrangement of the scene. For each object in a scene, depending on its properties, the robot may or may not be allowed to make contact with, tilt, or topple it. To ensure that these constraints are satisfied during non-prehensile interactions, a planner can query a physics-based simulator to evaluate the complex multi-body interactions caused by robot actions.
Unfortunately, it is infeasible to query the simulator for thousands of actions that need to be evaluated in a typical planning problem as each simulation is time-consuming. In this work, we show that (i) manipulation tasks \D{(specifically pick-and-place style tasks from a tabletop or a refrigerator)} can often be solved by restricting robot-object interactions to \textit{adaptive motion primitives} in a plan,
(ii) these actions can be incorporated as subgoals within a multi-heuristic search framework, and (iii) limiting interactions to these actions can help reduce the time spent querying the simulator during planning by up to $40\times$ in comparison to baseline algorithms. Our algorithm is evaluated in simulation and in the real-world on a PR2 robot using PyBullet as our physics-based simulator. Supplementary video: \url{https://youtu.be/ABQc7JbeJPM}.

\end{abstract}


\section{Introduction}\label{sec:intro}
Manipulation planning problems in domestic households, industrial manufacturing and warehouses require contact-rich interactions between a robot and the objects in the environment. As the amount of clutter in a scene increases, the likelihood of finding a completely collision-free trajectory for the manipulator decreases. This does not mean the task is impossible since we might still be able to complete it by moving the objects around. In these cases, non-prehensile interactions with objects can be much faster than deliberately rearranging the scene via a sequence of slow pick-and-place style prehensile maneuvers. In addition, each object in a cluttered scene is associated with constraints that define how a robot can manipulate it. For example, while we might be allowed to interact freely with a box of sugar, we might not be allowed to tilt or topple a cup of coffee.

We want to enable robots to grasp in clutter by using non-prehensile actions to interact with objects while satisfying any object-centric constraints, e.g. constraints that dictate whether or not we can make contact with, tilt, or topple an object. This domain of Manipulation Among Movable Obstacles (\MAMO)~\cite{StilmanMAMO} is derived from prior work on Navigation Among Movable Obstacles (\NAMO)~\cite{Wilfong91,NAMO}. Planning problems for \NAMO aim to find a feasible path between start and goal states for a mobile robot navigating in an environment with reconfigurable obstacles\footnote{\D{In our work `objects' may be movable, but `obstacles' are immovable.}}. \D{We focus on the class of \MAMO problems where the goal for the robot manipulator is a 6D pre-grasp pose of an object in $SE(3)$ without any constraints on the final poses of the movable objects. The task of planning the grasp itself is not addressed by our algorithm.}


\begin{figure}[t]
    \centering
    \includegraphics[width=0.95\columnwidth]{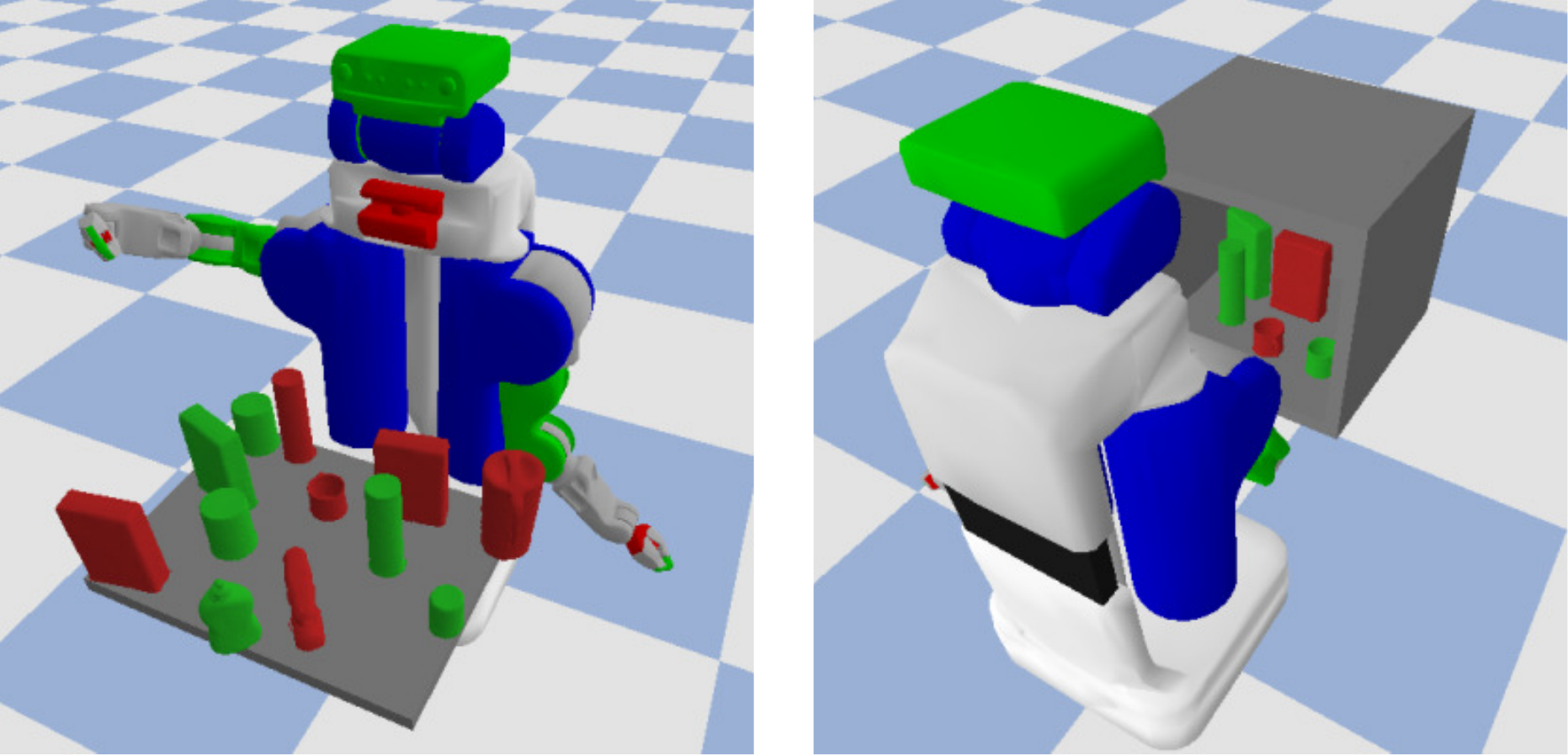}
    \caption{Manipulation tasks in cluttered tabletop (left) or refrigerator (right) workspaces require planners to account for complex multi-body interactions between the robot and objects (\textcolor{ObsGreen}{green} movable objects and \textcolor{ObsRed}{red} immovable obstacles). \D{The goal is to pickup a randomly selected immovable obstacle.}}
    \label{fig:mamosim}
\end{figure}


Motion planning for \MAMO is computationally costly because of two major challenges. First, we need to accurately model the dynamics of the robot-object and object-object interactions in the environment during planning. This requires the use of a high-fidelity physics-based simulator since hand-designed analytical models are hard to generalise for complex object geometries and cluttered scenes. The simulator is used to model the environment and predict the outcome of actions. The computational cost associated with running a simulator is high, which makes it infeasible to query the simulator for every action that needs to be evaluated. The second challenge is associated with the search space of the problem. Since the robot may interact with objects in the scene and reconfigure them, the search space needs to include the configuration space of all these objects. This makes the search space for a planning problem in this domain grow exponentially with the number of objects, and makes it computationally hard to find a solution.

In this work, we make an observation that many \MAMO problems can be solved effectively by restricting robot-object interactions to adaptive motion primitives and show how this observation can be exploited to structure an efficient search for a contact-rich motion. \textit{Adaptive motion primitives} ({\AMP}s, Section~\ref{subsec:amps}) are long-range actions generated on-the-fly such that they terminate in a valid goal state~\cite{CohenAMPs}. \D{In our domain, they are straight lines in the configuration space of the robot, between two states whose end-effector Cartesian coordinates are within $\delta$ of each other in Euclidean norm.
Since the goal in our domain is defined in the workspace of the robot, an \AMP is computed by linearly interpolating between a state that satisfies the $\delta$ threshold condition and an inverse kinematics (IK) solution of the goal pose.
}

\D{Our central assumption in this work limits robot-object interactions to the final action (an \AMP) in a plan. This restriction limits the class of \MAMO problems solvable by our algorithm to ones that require at most one robot action near the goal to make contact with the objects in the scene. We test our algorithm on random initialisations of the cluttered tabletop and refrigerator scenes from Fig.~\ref{fig:mamosim}. Empirically our results in Section~\ref{sec:exps} show that even with this restriction, our algorithm solves many \MAMO planning problems and is up to $40\times$ faster than competitive baselines in our experiments.}

Our main contributions towards robot manipulation planning among movable obstacles include:
\begin{itemize}
    \item \D{a \textit{two-stage planning approach} where we first parallely sample promising \AMP{}s, and then systematically use them in our planning algorithm to find a solution.}
    \item \D{the use of these \AMP{}s as \textit{subgoals} within a multi-heuristic search algorithm.}
    \item \D{an action evaluation scheme that minimises the time spent querying a simulator during planning.}
\end{itemize}

\section{Related Work}\label{sec:litreview}

The domain of Manipulation Among Movable Obstacles (\MAMO) is closely tied to prior work in the field of Navigation Among Movable Obstacles (\NAMO)~\cite{Wilfong91,NAMO}.
Past works in the \MAMO domain have taken one of two popular approaches - either solving problems via a sequence of pick-and-place style maneuvers~\cite{StilmanMAMO}, or limiting solutions to only planar robot-object interactions~\cite{BergSKLM08,DogarS12,King1}.
The rearrangement planning problem was extensively studied by King~\cite{KingPhD} in their thesis which focused on non-prehensile interactions.

In this work we use a physics-based simulator in-the-loop during planning to account for dynamics of robot-object and object-object interactions in the scene. Plaku et. al.~\cite{PlakuKV10} decompose the planning problem into a high-level discrete space, and a low-level sampling-based planner with a complex dynamics model (physics-based simulator). Zickler and Veloso~\cite{ZicklerV09} attempt to solve physics-based planning problems with the help of high-level, long-range robot behaviours. Dogar et. al.~\cite{DogarHCS12} simulate and cache multiple robot-object interactions, and use their result during planning to find feasible solutions. However, they do not allow any object-object interactions, which can be unavoidable in cluttered \MAMO scenes. \D{Similar to our work, the idea of planning till the proximity of the goal and using a more expensive specialized maneuver from within this proximity can be seen in \cite{grasprrt} in the context of grasp planning.}

Most relevant to our work in this paper are: a sampling-based planning algorithm \textsc{KPIECE}~\cite{KPIECE} and a search-based planning algorithm Selective Simulation~\cite{SelectiveSim}. Each uses different approaches to incorporate a simulator in-the-loop during planning. \textsc{KPIECE} is a sampling-based planner for applications with complex dynamics that uses an importance function over discretised cells of the robot workspace to guide exploration, but calls the simulator for all action evaluations. This is very computationally expensive for \MAMO and in our experimental comparisons with \textsc{KPIECE} we show that intelligently limiting the number of simulator queries can improve quantitative performance. Selective Simulation iteratively plans with simulations in a reduced search space (accounting for some objects) and executes the plan in a simulator (with all objects). If any object constraints are violated upon execution, it decides on one of these objects to be added to the search space for the next planning iteration. However, in the original paper Selective Simulation was evaluated for simple constraints of contact or toppling off the table. In addition to these, we consider constraints on how far obstacles can be tilted and how much velocity can be imparted to them. Selective Simulation is also prone to repeated simulations of similar actions which is time consuming and something we explicitly account for in our work with soft duplicate detection.
Our experimental analysis also includes comparison with Selective Simulation.




\section{Problem Formulation}\label{sec:problem}
\subsection{Search Space}\label{subsec:space}
In this work, we denote the robot manipulator as $\R$, and $\X_\R \subset \mathbb{R}^q$ as the configuration space for a $q$ degrees-of-freedom (dofs) manipulator. The set of objects in the scene is $\Obs = \{O_1, \ldots, O_n\}$, and the configuration of any object $\X_{O_i} \in SE(3)$ includes the 3D position and orientation. The robot is equipped with a set of actions $\A$.
The search space for a planning problem in the \MAMO domain is the Cartesian product of the robot and all object configuration spaces, denoted as $\X = \X_\R \times \X_{O_1} \times \cdots \times \X_{O_n}$.

\subsection{Object Constraints}\label{subsec:constraints}
Each object $O_i$ is associated with $m_i$ constraint functions $k_i^j: \X \rightarrow \{0, 1\}, 1 \leq j \leq m_i$ which return 1 if constraint $k_i^j$ is satisfied, 0 otherwise. For example, an \textit{immovable} obstacle (an object that is not allowed to be interacted with, for example, a wall) will contain a constraint function which evaluates to 1 so long as neither the robot nor any other object makes contact with it. We denote the evaluation of all constraints of an object as $K_i(x) = \left[k_i^1(x), \ldots, k_i^{m_i}(x)\right]$. If all constraints are satisfied, $K_i(x) = \boldsymbol{1}_{m_i}$.

We say a state is valid if all constraints for all objects are satisfied at that state\footnote{We omit the necessary robot kinematic and dynamic feasibility constraints from the definition of state validity for brevity.}. Formally, state $x$ is valid if $K_i(x) = \boldsymbol{1}_{m_i} \,\forall\, O_i \in \Obs, i \in \{1, \ldots, n\}$. We denote the space of valid states as $\X_V$.

\subsection{Problem Statement}\label{subsec:statement}
A planning problem in the \MAMO domain can be defined as the tuple $\PP = (\X, \A, x_S, \X_G, \T, c)$ where $x_S \in \X_V$ is the start state, $\X_G \subset \X, \X_G \cap \X_V \neq \emptyset$ is the set of goal configurations, $\T: \X \times \A \rightarrow \X$ is a deterministic transition function, and $c: \X \times \X \rightarrow \mathbb{R}_{\geq 0}$ is a state transition cost-function. We assume $\X_G$ is defined using a desired end-effector pose in $SE(3)$ \D{(a grasp location for an object; grasp planning is outside the scope of this problem)}, which leads to a set of possible manipulator configurations, some of which must be valid $(\X_G \cap \X_V \neq \emptyset)$.

A path $\pi$ of length $T$ is a sequence of states $\{x_1, \ldots, x_T\}$ and has cost $C(\pi) = \sum_{i=1}^{T-1} c(x_i, x_{i+1})$ where $x_{i+1} = \T(x_i, a_i)$. The goal for a \MAMO planning problem is to find a valid path from start to goal, i.e. a path made up of a sequence of valid states. Formally, we can write this as an optimisation problem:
\begin{equation}\label{eq:problem}
\begin{aligned}
    \text{find } &\pi^* = \argmin_\pi C(\pi) \\
    \text{s.t. } &x \in \X_V, \,\forall\, x \in \pi &&\text{(path of valid states)} \nonumber \\
    &x_1 = x_S, x_T \in \X_G &&\text{(start, goal constraints)} \nonumber \\
    &x_{i+1} = \T(x_i, a_i), a_i \in \A, \!\!\!\!\!&&\forall x_i, x_{i+1} \in \pi \nonumber \\
    & &&\text{(transition dynamics)}
\end{aligned}
\end{equation}


\subsection{Graph Representation}\label{subsec:graph}
\D{
We solve \MAMO planning problems using a search-based planning algorithm in $\X$. Our graph representation contains two types of actions in $\A$ - \textit{simple primitives}
and \textit{adaptive motion primitives} (\AMP{}s). Each simple primitive changes one joint angle of a robot by a small amount\footnote{\D{In our implementation, $4^\circ$ or $7^\circ$ depending on the joint in our implementation.}}. In comparison, an \AMP is computed on the fly and can change all coordinates in $\X_\R$. A consequence of our core assumption (Section~\ref{subsec:assumptions}) is that for simple primitives $a_\text{s} \in \A$, a valid transition $x^\prime = \T(x, a_\text{s})$ implies that the state $x$ and the successor state $x^\prime$ only differ in the robot configuration (in $\X_\R$). For \AMP{}s $a_\AMP \in \A$, a valid transition $x^\prime = \T(x, a_\AMP)$ can lead to differences in object configurations ($\X_{O_1} \times \cdots \times \X_{O_n}$) in addition to a difference in $\X_\R$.
}


\section{Approach}\label{sec:approach}


\subsection{Action Evaluation}\label{subsec:actions}
Following the ideas outlined in~\cite{SelectiveSim}, we decompose our action evaluation scheme into a relatively fast collision checking routine and a much slower physics-based simulation. Collision checking involves checking for volumetric overlaps between the collision models of the robot and objects. This is computationally a relatively cheap operation that can be easily implemented with the use of a distance field. If and when necessary, an action that passes this collision checking phase might need to be simulated to determine whether or not it violates any object constraints.

The set of objects $\Obs$ in a scene can be separated into two subsets - \textit{movable} objects $\Obs_M$ that the robot is allowed to interact with, and \textit{immovable} obstacles $\Obs_I = \Obs\setminus\Obs_M$. We assume that this separation is known a priori.

\begin{definition}[Phase 1 validity]
We say an action $a \in \A$ from state $x \in \X_V$ is \textit{Phase 1 valid} if it does not make contact with any immovable obstacle $O \in \Obs_I$.
\end{definition}

\begin{definition}[Phase 2 validity]
We say an action $a \in \A$ from state $x \in \X_V$ is \textit{Phase 2 valid} if it is Phase 1 valid and it does not result in any object constraint violations.
\end{definition}

The fast collision checking routine is used to determine Phase 1 validity of an action as it can quickly detect overlaps with immovable obstacles. This can also determine Phase 2 validity if there is no overlap with any object. Note that for $a \in \A,\, x \in \X_V,\, x^\prime = \T(x, a)$, we call the collision checking routine for all intermediate states between $x$ and $x^\prime$, including $x^\prime$ but not $x$. In the case when an action is Phase 1 valid, but also makes contact with some movable object(s), Phase 2 validity can only be determined after simulating it. This is because we need to account for the complex multi-body interactions that might result upon executing the action. These interactions might violate object constraints due to a movable object-immovable obstacle contact, or the robot violating other movable object constraints such as tilting or toppling. We emphasise that determining Phase 1 validity of an action is computationally much cheaper (around $30 \si{\milli\second}$ per action evaluation for a 7 dof manipulator) than determining Phase 2 validity which requires simulating it (e.g., around $1.5\si{\second}$ per action of a 7 dof manipulator in PyBullet).

\subsection{Adaptive Motion Primitives}\label{subsec:amps}

\begin{figure}[t]
    \centering
    \includegraphics[width=0.45\columnwidth]{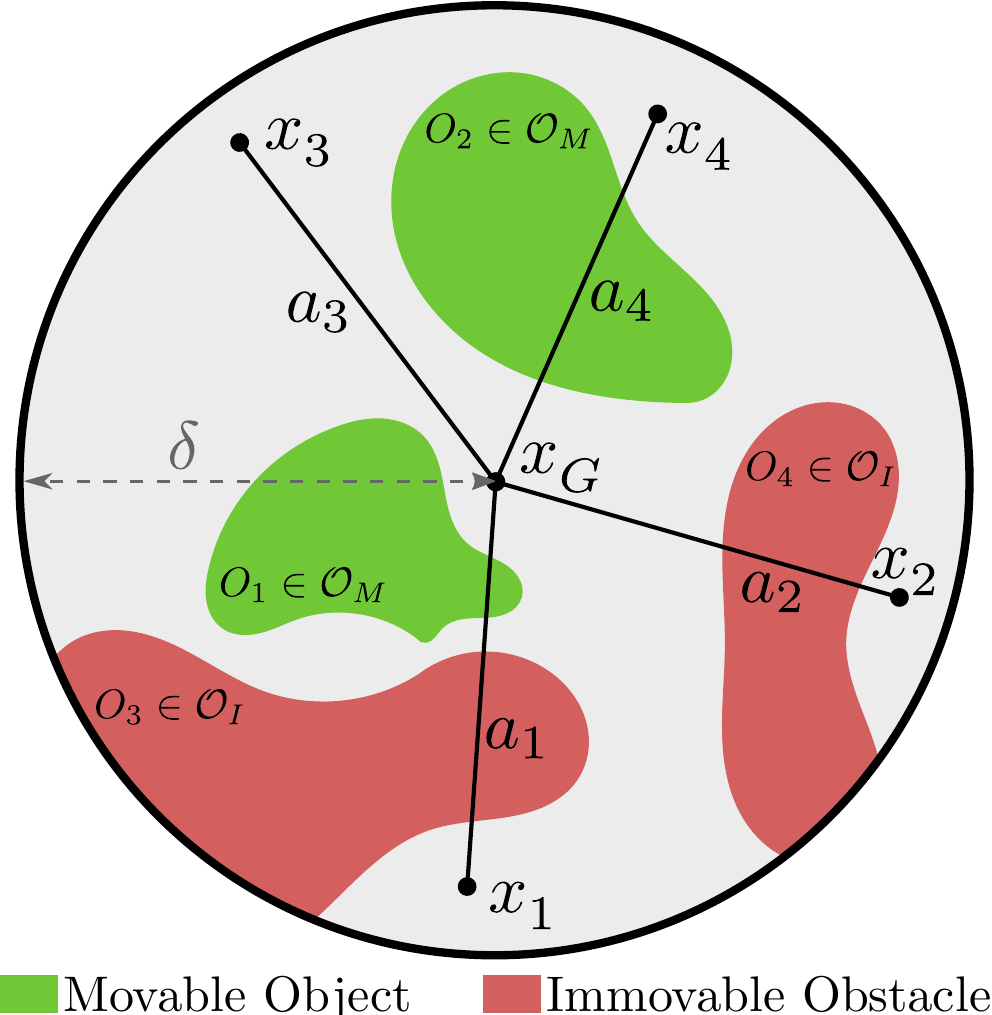}
    \caption{For a particular goal state configuration $x_G \in \X_\R$, we can generate \AMP{}s from several states $x_i$ within distance $\delta$ from it. This $\delta$-sphere in configuration space $\X_\R$ might be occupied by both movable (\textcolor{ObsGreen}{green}) objects and immovable (\textcolor{ObsRed}{red}) obstacles. This can lead to invalid actions $a_1, a_2$, valid action $a_3$, and Phase 1 valid action $a_4$ whose Phase 2 validity will be determined after simulation.}
    \label{fig:amps}
\end{figure}

Adaptive motion primitives (\AMP{}s) are IK-based motion primitives that are generated on-the-fly as part of our algorithm~\cite{CohenAMPs} and included in $\A$. \D{For any robot configuration $x \in \X_\R$, if the robot's 3D Cartesian end-effector pose is within $\delta = 0.2\si{\meter}$ of the goal end-effector pose $\X_G$, we generate an \AMP that tries to connect $x$ to $\X_G$. This is done by obtaining an IK solution $x_G \in \X_\R$ for $\X_G$, and linearly interpolating between $x$ and $x_G$. We choose this value of $\delta$ based on basic domain knowledge like the size of our workspaces and obstacles therein. We did not tune this value to improve performance.}

\D{The validity of an \AMP (Phase 1 or Phase 2) is dependent on checking all interpolated states between $x \text{ and } x_G$.} Fig.~\ref{fig:amps} shows our action evaluation strategy from Section~\ref{subsec:actions} for \AMP{}s. Since \AMP{}s terminate in a goal state, they can only be included in valid paths as the final action.

\subsection{Assumptions}\label{subsec:assumptions}

We make one assumption for solving \MAMO planning problems in this work which is closely related to \AMP{}s and their use in our search algorithm. In this subsection we hope to provide an intuitive justification for this assumption.

\begin{assumption}\label{assume:amps}
We only need to simulate \AMP{}s to find a valid solution for a \MAMO planning problem.
\end{assumption}

For grasping and reaching in cluttered scenes like those we consider in this work, there is a large volume of object-free space between the start configuration and goal region. Interactions with objects are necessary when the robot is in a region with a high degree of clutter, and clutter near the goal is often the most pertinent for finding a feasible plan.
\D{in the tabletop and refrigerator workspaces in our work}.
Based on this observation we delay interacting with objects until the robot reaches a configuration near the goal.

\begin{figure}[t]
    \centering
    \includegraphics[width=0.85\columnwidth]{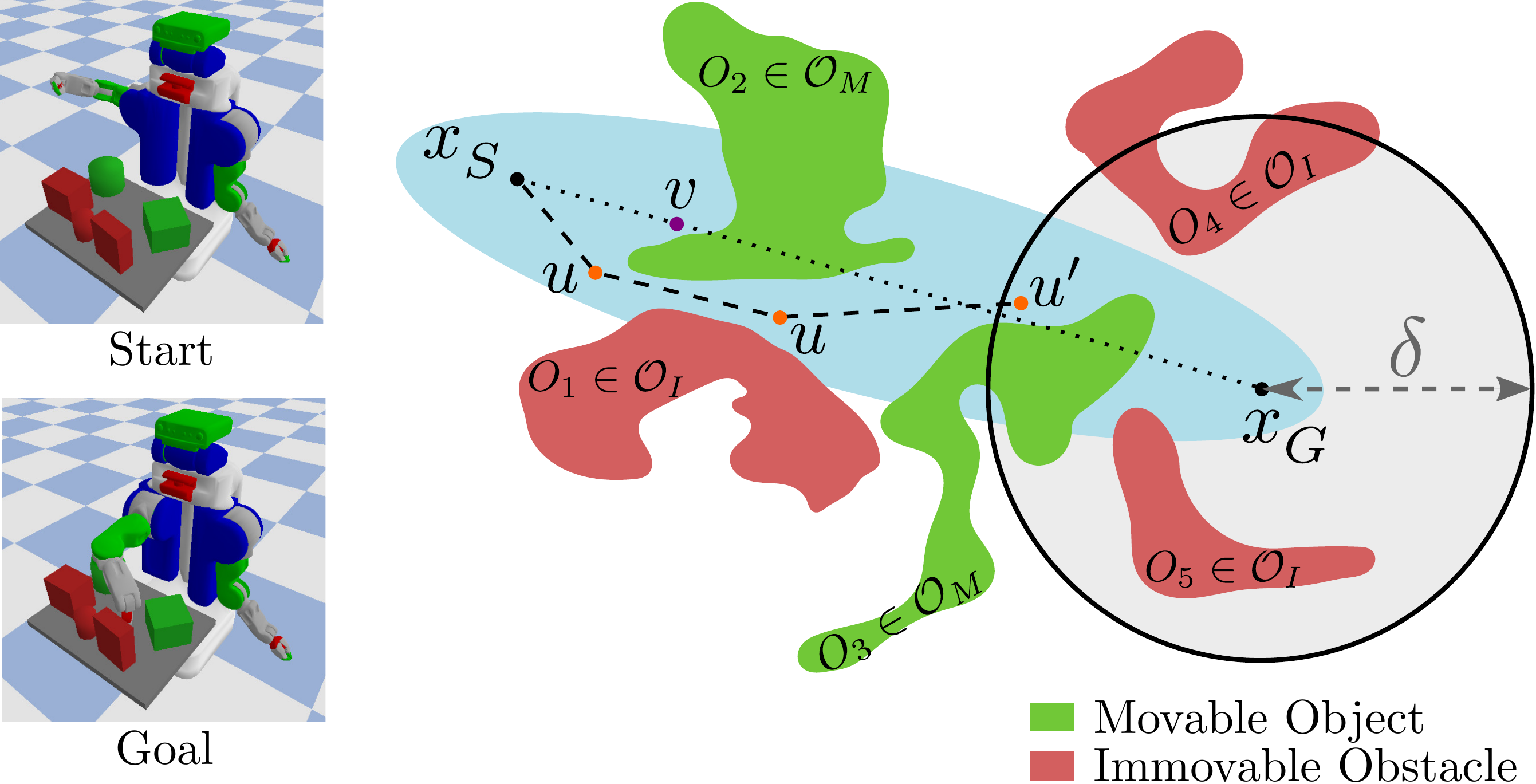}
    \caption{Consider planning between start and goal configurations shown on the left. A ``greedy'' shortest-path search algorithm in the \MAMO domain would proceed along the dotted path, exploring states in the \textcolor{EllipseBlue}{light blue} region, and spend a lot of time simulating interactions with object $O_2$ near the \textcolor{CirclePurple}{purple} state $v$. Due to the assumptions we make, our search algorithm \textsc{SPAMP} proceeds along the dashed path via \textcolor{CircleOrange}{orange} states $u$, and only starts simulating \AMP{}s from states beyond $u^\prime$ in the $\delta$-sphere around goal $x_G$.}
    \label{fig:assumptions}
\end{figure}

For a preset value of $\delta$, we restrict robot-object interactions until the end-effector is within $\delta$ of the goal pose. Since \AMP{}s are long-range actions contained inside this $\delta$-sphere, interactions that are vital to the success of a plan are often the terminal \AMP{}s in a plan (\D{in comparison to the short-range simple primitives which do not lead to meaningful interactions}). This leads us to Assumption~\ref{assume:amps}. Consequently, since \AMP{}s are terminal actions, the valid solution paths we find only contain a single action which interacts with obstacles. It is important to note that restricting interactions to a single action does not limit the number of objects the robot can interact with. We illustrate the effect of Assumption~\ref{assume:amps} in Fig.~\ref{fig:assumptions} by comparing our algorithm against a naive search algorithm.

\begin{figure}[t]
    \centering
    \includegraphics[width=0.7\columnwidth]{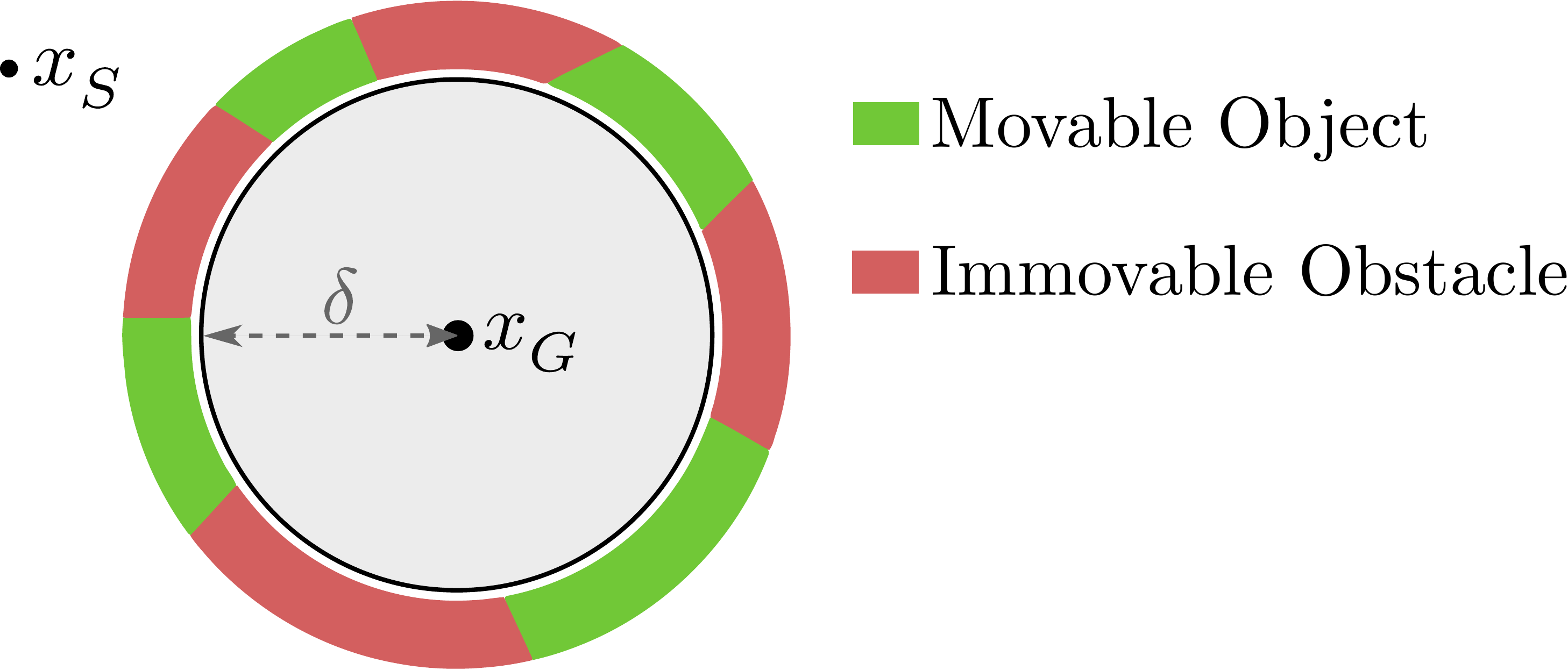}
    \caption{In case there is no interaction-free path between start state $x_S$ and the $\delta$-sphere around any goal state $x_G$, our assumption of only simulating terminal \AMP{}s will cause our algorithm to not return a solution. \D{A greater value of $\delta$ in this case could make this problem solvable by our algorithm}.}
    \label{fig:failure}
\end{figure}

\D{This assumption restricts the space of \MAMO planning problems solvable by our algorithm to those that require at most one \AMP to interact with objects near the goal configuration. Our success rates from Section~\ref{sec:exps} suggest that this assumption is not restrictive for the high-clutter scenes we consider in this work (Fig.~\ref{fig:mamosim}).} In cases where no such configuration near the goal is achievable by the manipulator as shown in Fig.~\ref{fig:failure}, our algorithm will fail to find a solution.
\D{It might still be possible to find solutions in these cases by dynamically changing $\delta$ to find a valid \AMP, but we have not explored this yet.}

Assumption~\ref{assume:amps} helps us deal with the two major computational challenges for \MAMO:
\begin{enumerate}
    \item The number of calls to the simulator go down significantly, as we now only simulate terminal \AMP{}s that interact with movable objects as opposed to any action that interacts with movable objects.
    \item Since only terminal \AMP{}s might be simulated, we can plan from $x_S$ to inside the $\delta$-sphere around some goal configuration purely in $\X_\R$. This means that the search space for finding a path from $x_S$ to $x_{T-1}$ reduces from $\X = \X_\R \times \X_{O_1} \times \cdots \times \X_{O_n}$ to $\X_\R$, thereby tackling the issue of a prohibitively large search space in cluttered environments.
\end{enumerate}


\subsection{Subgoals}\label{subsec:subgoals}
We solve \MAMO planning problems using a search-based planning algorithm. The performance of these algorithms is dependent on the quality of the heuristic functions used. As it is often infeasible to create a single heuristic that can perfectly guide the search from start to goal in all scenarios, it is common practice in high-dimensional spaces to use a multi-heuristic framework. Multiple heuristics guide the search along multiple promising directions, which can help overcome local minima associated with any one heuristic.

Given the fact that simulations are the computational bottleneck in our domain, and the assumption that we only simulate \AMP{}s, it is helpful to guide a search-based planning algorithm to regions of the search space where a valid \AMP likely exists.
\D{Our two-stage planning approach discussed in Section~\ref{sec:algo} first finds \AMP{}s that are Phase 1 or Phase 2 valid}, and generates heuristic functions that guide the search to the beginning of these actions. For an \AMP from state $x_{T-1} \in \X_V$, the corresponding heuristic function is
a simple Euclidean distance from $x_{T-1}$ in $\X_\R$. \D{This first stage is run in parallel across multiple simulator instances, one per \AMP sampled. The second stage runs a multi-heuristic search to find a path from start state $x_S$ to a goal state in $\X_G$.}

The beginning of the \AMP $x_{T-1}$ can be thought of as a \textit{subgoal} for the planner since we guide the search towards it. While it is not necessary to reach the subgoal,
exploring the search space near it can help find a solution.

\D{If an \AMP $a_\AMP$ is Phase 2 valid, and the subgoal $x_{T-1}$ is reachable without making contact with any object, we do not need to simulate actions on the way to $x_{T-1}$. It suffices to find a collision-free path in $\X_\R$ from start $x_S$ to $x_{T-1}$, and append $x_T = \T(x_{T-1}, a_\AMP)$ for a valid \MAMO solution.}

If we only have access to Phase 1 valid subgoals, we allow our planning algorithm to simulate any Phase 1 valid \AMP{}s that are generated during the search. To avoid the rare case when a subgoal is Phase 2 valid and unreachable (\D{perhaps due to kinematic limits or a scenario like Fig.~\ref{fig:failure}}), we allow our planner to simulate Phase 1 valid \AMP{}s after time $t$ has elapsed during planning (Algorithm~\ref{alg:highlevel}, Line~\ref{line:t})\footnote{\D{We set $t = 30 \si{\second}$ for our experiments.}}. This is implemented by maintaining a priority queue of all Phase 1 valid \AMP{}s generated by the search, and simulating them in order after time $t$.





\subsection{Soft Duplicate Detection for Action Evaluation}\label{subsec:edgeeval}

\begin{figure}[t]
    \centering
    \includegraphics[width=0.4\columnwidth]{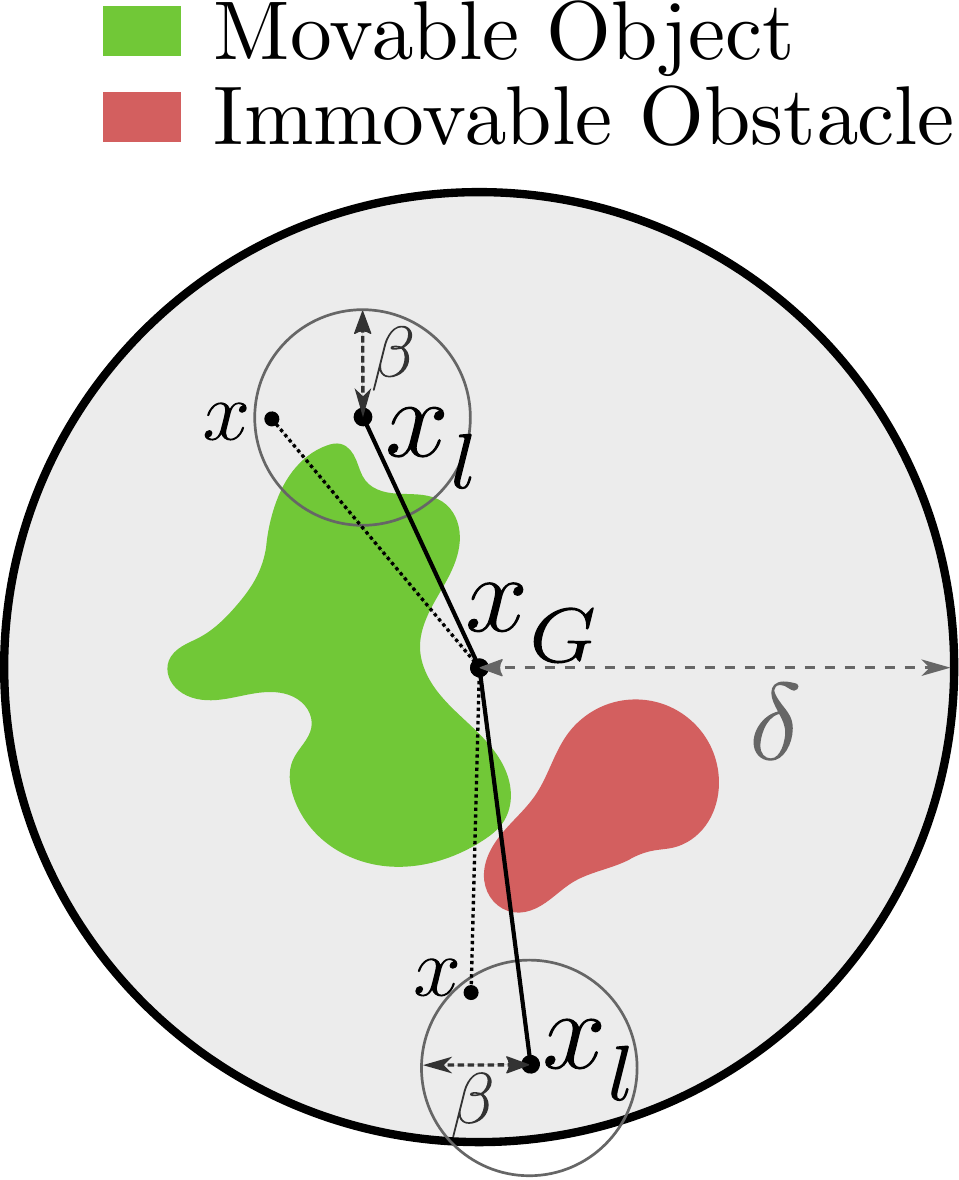}
    \caption{If an \AMP from any $x_l \in L$ is found to be invalid, we postpone the simulation of \AMP{}s from any other state $x$ that is within $\beta$ distance from it by inflating the heuristic value of that action.}
    \label{fig:edgeeval}
\end{figure}

Assumption~\ref{assume:amps} states that we only simulate Phase 1 valid \AMP{}s during planning. However, since the number of such \AMP{}s in cluttered environments can be very large, we further optimise the calls to the simulator by employing a soft duplicate action detection scheme~\cite{WeiEdgeEval}.

Soft duplicate detection estimates the similarity between an action that needs to be evaluated in simulation and an action that has already been simulated and deemed invalid (one which violated object constraints). If they are very similar then it is likely that the new action would also violate some of the same constraints and be invalid. \D{The idea of preferring promising actions based on the validity of similar actions can also be found in \cite{hernandez2019lazy}.}

We maintain a list of states $L$ from which an \AMP has been simulated and found to be invalid. For any new state $x \in \X_V$ from which we find a Phase 1 valid \AMP, we first compute its Euclidean distances in $\X_\R$ to states in $L$. Given a preset threshold $\beta$, if $\lVert x - x_l \lVert_2 < \beta$ for any $x_l \in L$ we postpone the simulation of the \AMP from $x$ by artificially inflating the heuristic value of $x$ and re-inserting it into the priority queue it was expanded from.
This ensures that $x$ might be re-expanded from the same priority queue at a later time, at which point we would simulate the \AMP from it to a goal configuration. Fig.~\ref{fig:edgeeval} shows how this process implicitly creates $\beta$-spheres in configuration space around states that lead to invalid \AMP{}s due to any constraint violation.



\section{Algorithm}\label{sec:algo}

\begin{algorithm}[t]
\begin{small}
\caption{\small{Simulation-based Planning with \AMP{}s (\textsc{SPAMP})}}\label{alg:highlevel}
\textbf{Input:} Planning problem $\PP$, number of \AMP subgoals $N$, number of \AMP samples $M$, simulation start time $t_\text{sim}$, planning timeout $t_\text{max}$ \\
\textbf{Output:} solution path $\pi$
\begin{algorithmic}[1]
\Procedure{SPAMP}{$\PP, N, M, t_\text{sim}, t_\text{max}$}
    \State $H \gets \texttt{GetValidSubgoals}(N, M)$ \label{line:sample}
    \State \Comment{Phase 1 or Phase 2 valid subgoals.}
    \State $t \gets t_\text{sim}$ \Comment{Simulations are allowed starting from time $t$.} \label{line:t}
    \If{$\lvert H \lvert = 0$ or $\lvert \texttt{IsPhase2Valid}(H) \lvert = 0$}
        \State $t \gets 0$ \Comment{$\lvert\cdot\lvert$ is the set cardinality operation.}
    \EndIf
    \State $OPEN \gets \texttt{InitialiseHeuristics}(H)$
    \State $\pi \gets \textsc{Plan}(\PP, H, OPEN, t, t_\text{max})$ \label{line:plan}
    \Return $\pi$
\EndProcedure

\end{algorithmic}
\end{small}
\end{algorithm}

Algorithm~\ref{alg:highlevel} is a high-level overview of our two-stage planning pipeline. We call our algorithm Simulation-based Planning with \AMP{}s (\textsc{SPAMP}). The \textsc{GetValidSubgoals} subroutine in Line~\ref{line:sample} selects subgoals via rejection sampling. $M$ Phase 1 valid \AMP{}s are randomly sampled and simulated in parallel. $N$ Phase 2 valid subgoals are returned (if available), else a combination of Phase 1 and Phase 2 valid make up the $N$ returned subgoals (first stage). This section goes into more details about our specific planning algorithm from Line~\ref{line:plan} of Algorithm~\ref{alg:highlevel} (second stage).



\subsection{Multi-Heuristic Framework for \MAMO}\label{subsec:mha}


We use Multi-Heuristic A* (MHA*)~\cite{mha*} as our search algorithm in this work. MHA* maintains multiple priority queues, one for each heuristic that is used. It was originally developed under the assumption that all action evaluations take roughly the same amount of time. This meant priority queues could be selected round robin for state expansions to equitably distribute computational resources across the queues. However, our action evaluations have varying time complexity (checking for Phase 1 vs. Phase 2 validity).
Since some queues might need many simulator calls, and some queues might never query the simulator, a simple round robin strategy would lead to an uneven distribution of computational resources across the queues. For this reason, we prioritise state expansions from queues that the search has spent the least time expanding states from thus far. This time-based prioritisation of queues in MHA* leads to a much more equitable allocation of computational resources for \MAMO.


\subsection{Planning Algorithm}\label{subsec:algo}
Algorithm~\ref{alg:details} contains details from the preceding sections to provide a more in-depth look at our planning algorithm. \D{Details of the MHA* planning algorithm can be found in the original publication~\cite{mha*}}. In our algorithm, $OPEN$ is a set of priority queues, each of which is defined by a heuristic function. Following standard MHA* terminology, our \textit{anchor} heuristic is a 3D Breadth-First Search (BFS) heuristic computed from the specified Cartesian goal end-effector position~\cite{CohenSMPL}, \D{taking into account the immovable obstacles in the scene}. Every other heuristic is defined by a corresponding \AMP subgoal as per Section~\ref{subsec:subgoals}. The action set $\A$ is made up of the \D{simple primitives and \AMP{}s via the function \texttt{GenerateAMP}.}
\D{Simple primitives and \AMP{}s are denoted by $a_\text{s}$ and $a_\AMP$ respectively (Section~\ref{subsec:graph}).}

\D{\textsc{SPAMP} terminates if the next-best state to expand $x$ is in the goal set or we have already found a Phase 2 valid \AMP from it. For every other $x$, the \textsc{Expand} function generates and evaluates all possible successor states of $x$. These necessarily include successors $x^\prime = \T(x, a_\text{s}) \,\forall\, a_\text{s} \in \A$ (Line~\ref{line:prims}).
In addition, we may generate and evaluate an \AMP from $x$ to $x_G \in \X_\R$, an inverse-kinematics solution for $\X_G$ (Line~\ref{line:ik}). This evaluation (Line~\ref{line:genamp} onwards) occurs if $x$ passes the end-effector distance check (Line~\ref{line:eecheck}) and soft duplicate check (Line~\ref{line:duplicate}). The \texttt{IsInteraction} function returns true if $a_\AMP$ `collides' with a movable object during the Phase 1 validity check (Line~\ref{line:phase1}).}



\begin{algorithm}[t]
\begin{small}
\caption{\textsc{SPAMP}}\label{alg:details}
\begin{algorithmic}[1]
\Procedure{Plan}{$\PP, H, OPEN, t, t_\text{max})$}
    \State $\texttt{Insert}(OPEN, x_S)$ \Comment{Add to all queues.}
    \While{$OPEN$ is not empty}
        \State $h \gets \texttt{BestQueue}(OPEN)$ \Comment{Time-based selection.}
        \State $x \gets \texttt{BestState}(h)$ \Comment{Pop from all queues.} \label{line:best}
        \If{$x \in \X_G$}
            \Return $\texttt{ExtractPath}(x)$
        \EndIf
        \If{$\exists$ Phase 2 valid \AMP from $x \in H$}
            \State $a_\AMP \gets$ Phase 2 valid \AMP from $x \in H$
            \Return $\texttt{ExtractPath}(x) \cup \{\T(x, a_\AMP)\}$
        \EndIf
        \State $\textsc{Expand}(x, OPEN, t)$ \label{line:expand}
    \EndWhile
\EndProcedure

\Procedure{Expand}{$x, OPEN, t$}
    \For{$a_\text{s} \in \A$} \Comment{Simple primitives only.} \label{line:prims}
        \State $x^\prime \gets \T(x, a)$.
        \If{$x^\prime \in \X_V$}
            \State $\texttt{Insert}(OPEN, x^\prime)$ \label{line:insert0}
        \EndIf
    \EndFor
    \If{\D{$\lVert \texttt{FK}(x) - \X_G \lVert_2 \leq \delta$}}\label{line:eecheck} \Comment{\D{3D end-effector poses.}}
        \If{$x$ has soft duplicate} \label{line:duplicate}
            \State $\texttt{InflateHeuristic}(x)$
            \State $\texttt{Insert}(OPEN, x)$ \label{line:insert1}
            \Return
        \EndIf
        \State \D{$x_G \gets \texttt{IK}(\X_G)$}\label{line:ik} \Comment{\D{Inverse kinematics.}}
        \State $a_\AMP \gets \texttt{GenerateAMP}(x, x_G)$ \label{line:genamp}
        \If{$\texttt{IsPhase1Valid}(a_\AMP)$} \label{line:phase1}
            \If{$\texttt{IsInteraction}(a_\AMP) \text{ and } t_\text{elapsed} > t$} \label{line:interaction}
                \State $x^\prime \gets \texttt{Simulate}(a_\AMP)$ \label{line:simulate}
                \If{$x^\prime \in \X_V$} \label{line:phase2}
                    \State $\texttt{Insert}(OPEN, x^\prime)$ \label{line:insert2}
                \Else
                    \State $\texttt{Insert}(L, x)$
                \EndIf
            \ElsIf{not $\texttt{Interaction}(a_\AMP)$} \label{line:apply}
                \State $x^\prime \gets \T(x, a_\AMP)$.
                \State $\texttt{Insert}(OPEN, x^\prime)$ \label{line:insert3}
            \EndIf
        \EndIf
    \EndIf

\EndProcedure

\end{algorithmic}
\end{small}
\end{algorithm}

\section{Experimental Results}\label{sec:exps}

We run all our experiments on the PR2 robot and use PyBullet~\cite{coumans2019} as our physics-based simulator. We run experiments in two different workspaces - a tabletop and a refrigerator. The objects in a scene are divided into immovable and movable subsets prior to planning.
The robot is allowed to interact with the movable objects but cannot tilt them excessively, cause them to fall over or outside the workspace (table or refrigerator), or impart high velocities. Neither the robot nor any movable object can make contact with immovable obstacles.


\subsection{Comparative Quantitative Evaluation in Simulation}\label{subsec:sim}

\begin{table*}[t]
\caption{Simulation study of various planners in the \MAMO domain.}
\label{tab:exps_sim}
\begingroup
\setlength{\tabcolsep}{3pt}
\centering
\scriptsize
\begin{tabular}{ll*{6}c*{6}c}
\toprule
& & \multicolumn{12}{c}{\footnotesize\textbf{Planning Algorithms}} \\
\cmidrule{3-14}
& & \multicolumn{6}{c}{\footnotesize Tabletop Workspace (6 movable, 6 immovable)} & \multicolumn{6}{c}{\footnotesize Refrigerator Workspace (3 movable, 2 immovable)}\\
\cmidrule(lr){3-8}\cmidrule(lr){9-14}
\footnotesize\textbf{Metrics} & \footnotesize\textbf{Scenario} & \footnotesize\textbf{\textsc{SPAMP}} & \footnotesize\textsc{K1} & \footnotesize\textsc{K2} & \footnotesize\textsc{K3} & \footnotesize\textsc{SS} & \footnotesize\textsc{SS2} & \footnotesize\textbf{\textsc{SPAMP}} & \footnotesize\textsc{K1} & \footnotesize\textsc{K2} & \footnotesize\textsc{K3} & \footnotesize\textsc{SS} & \footnotesize\textsc{SS2} \\ \midrule 
Success $\%$ & Overall & \cellcolor{PaleGreen1}\textbf{99$\%$} & 91$\%$ & 92$\%$ & 85$\%$ & 87$\%$ & 91$\%$ & 93$\%$ & 51$\%$ & 38$\%$ & \cellcolor{PaleGreen1}\textbf{94$\%$} & 87$\%$ & 91$\%$ \\ \midrule 
\multirow{2}{*}{\shortstack{Planning\\ Time (\si{\second})}} & Easy & \cellcolor{PaleGreen1}\textbf{5 $\pm$ 5} & 267 $\pm$ 301 & 339 $\pm$ 351 & 211 $\pm$ 369 & 6 $\pm$ 20 & 8 $\pm$ 26 & \cellcolor{PaleGreen1}\textbf{3 $\pm$ 3} & 116 $\pm$ 293 & 210 $\pm$ 443 & 167 $\pm$ 222 & 7 $\pm$ 22 & 14 $\pm$ 57 \\
& Difficult & \cellcolor{PaleGreen1}\textbf{11 $\pm$ 16} & 318 $\pm$ 310 & 464 $\pm$ 416 & 210 $\pm$ 327 & 22 $\pm$ 41 & 38 $\pm$ 55 & \cellcolor{PaleGreen1}\textbf{8 $\pm$ 13} & 58 $\pm$ 119 & 128 $\pm$ 335 & 249 $\pm$ 425 & 57 $\pm$ 126 & 63 $\pm$ 157 \\ \midrule 
\multirow{2}{*}{\shortstack{Simulation\\ Time (\si{\second})}} & Easy & \cellcolor{PaleGreen1}\textbf{0 $\pm$ 0} & 267 $\pm$ 301 & 339 $\pm$ 351 & 42 $\pm$ 88 & 4 $\pm$ 18 & 4 $\pm$ 14 & \cellcolor{PaleGreen1}\textbf{0 $\pm$ 0} & 116 $\pm$ 293 & 208 $\pm$ 441 & 50 $\pm$ 89 & 3 $\pm$ 18 & 8 $\pm$ 35 \\
& Difficult & \cellcolor{PaleGreen1}\textbf{1 $\pm$ 7} & 318 $\pm$ 310 & 463 $\pm$ 415 & 90 $\pm$ 131 & 4 $\pm$ 14 & 16 $\pm$ 31 & \cellcolor{PaleGreen1}\textbf{1 $\pm$ 6} & 57 $\pm$ 118 & 137 $\pm$ 334 & 44 $\pm$ 92 & 20 $\pm$ 82 & 24 $\pm$ 62 \\ \bottomrule
\end{tabular}
\endgroup
\end{table*}

We compare the performance of our algorithm Simulation-based Planning with \AMP{}s (\textsc{SPAMP}) against relevant state-of-the-art baseline algorithms that can be applied to the \MAMO domain - \textsc{KPIECE}~\cite{KPIECE} and Selective Simulation (\textsc{SS})~\cite{SelectiveSim}, which were described in Section~\ref{sec:litreview}. \D{Three variants of \textsc{KPIECE} were tested. The first two use use the \textsc{KPIECE} implementation from OMPL~\cite{sucan2012the-open-motion-planning-library}, but differ in how goal biasing is implemented. \textsc{K1} precomputes a set of 3 valid goal configurations by running IK before planning. \textsc{K2} runs IK online (with random seeds) every time the search tree is grown towards the goal. \textsc{K3} is our implementation of \textsc{KPIECE}\footnote{\D{Per~\cite{KPIECE}, we implement multiple levels of discretisation, goal biasing, and add all intermediate states of `motions' to the search tree (something OMPL does not do). Additionally, we only simulate Phase 1 valid motions.}}. The projective space for all \textsc{KPIECE} algorithms is the 3D Cartesian coordinate for the end-effector (via robot forward kinematics). In addition, we implemented a modified version of Selective Simulation (\textsc{SS2}) which includes soft duplicate detection on top of \textsc{SS}.}


A planning problem is initialised with objects selected at random from the YCB Object Dataset~\cite{YCB}. \D{Objects are placed in the workspace at random. The goal for the PR2 is to reach a pre-grasp pose for an immovable obstacle. Object masses are obtained from the YCB dataset, and their PyBullet friction coefficients are randomly sampled from the interval $[0.5, 1.1]$ as the dataset does not provide any friction coefficient values. We run all planners on 180 randomly initialised planning problems in both workspaces.}

Table~\ref{tab:exps_sim} shows quantitative results for all planners. We use 12 objects on the tabletop (6 movable and 6 immovable), and 5 objects in the refrigerator (3 movable and 2 immovable)\footnote{\D{The tabletop is $0.6\si{\meter}\times0.8\si{\meter}$, and refrigerator is $0.6\si{\meter}\times0.6\si{\meter}\times0.6\si{\meter}$.}}. Sample initialisations of these workspaces are shown in Fig.~\ref{fig:mamosim}.
\textsc{SPAMP} uses $N = 3$ subgoals from $M = 8$ samples in \textsc{GetValidSubgoals}. All planners were given a maximum planning time of $1800 \si{\second}$. We divide all planning problems into two scenarios based on how long it takes a \textsc{Naive} planner (a vanilla MHA* algorithm with only the 3D BFS heuristic to the goal) to solve them. \textsc{Naive} simulates all Phase 1 valid \AMP{}s. Problems solved by \textsc{Naive} in less than $100 \si{\second}$ are `Easy', and the rest are `Difficult'.

\textsc{SPAMP} achieves the highest success rate across all algorithms which shows that our assumption from Section~\ref{subsec:assumptions} is not too restrictive for the \MAMO domain. In terms of planning times, \textsc{SPAMP} is $30-50\times$ faster than \textsc{KPIECE} and \textsc{KPIECE2}, and $2-8\times$ faster than \textsc{SelSim} and \textsc{SelSim2}. \textsc{SelSim} is most competitive in terms of planning times, but it is still $2-7\times$ slower than \textsc{SPAMP} for difficult planning problems. By design, \textsc{SPAMP} spends most of the simulation time finding valid subgoals for the search which is still comparable to \textsc{SelSim}, and one or two orders of magnitude less than the other three baseline algorithms.

\D{\textsc{KPIECE} by default simulates all actions in the search tree, spends almost all of its planning time in simulation, and in OMPL samples a random state in $\X_\R$ $95\%$ of the time ($\X_\R \subset \mathbb{R}^7$ for a PR2 arm). Since each simulation takes around $1.5\si{\second}$, and mostly random point-to-point exploration of $\X_\R$ (as implemented in OMPL) is wasteful, planning times grow quickly with the number of actions \textsc{KPIECE} evaluates.}


\subsection{Runs on a Physical Robot}\label{subsec:realworld}

We setup the tabletop workspace experiment with the PR2 robot in our laboratory. We also set up a rudimentary experiment to calculate the coefficient of static friction as the tangent of the incline angle of the table at which the objects start sliding. We used this friction coefficient in our simulator in an attempt to minimise the sim-to-real gap. We selected 6 objects at random to initialise our scene and used a search-based object localisation algorithm~\cite{Agarwal2020PERCH2} on an NVidia TITAN X GPU to detect the object poses for simulator initialisation. \textsc{SPAMP} was given a $30 \si{\second}$ planning timeout, and objects were instantiated in the simulator as movable with $75\%$ probability.

The quantitative data from execution of 39 plans on the physical robot is shown in Table~\ref{tab:realworld}. A success rate of $82\%$ means that in 7 out of 39 of the executions, the robot violated an obstacle constraint in the real-world. Since the plan found by the robot must have been valid in simulation, constraint violations in the real-world are due to a mismatch between the simulator and the real-world. This could occur due to modeling errors for the obstacles, execution errors on the robot, or perception errors in object localisation.

\begin{figure}[t]
    \centering
    \includegraphics[width=0.5\columnwidth]{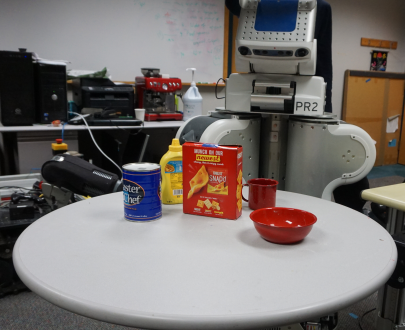}
    \caption{Experimental setup for a PR2 robot in front of a tabletop workspace for \MAMO.}
    \label{fig:table}
\end{figure}

A qualitative assessment of the 39 executions indicates that the two main sources of error in our experiment were inaccurate friction coefficients and object localisations. Fig.~\ref{fig:table} shows an image of our experimental setup. Our supplemental video submission also includes successful executions by the PR2 in a refrigerator and cabinet workspace.

\begin{table}[]
\caption{Quantitative Performance for Real-World Experiments}
\label{tab:realworld}
\begingroup
\setlength{\tabcolsep}{4pt}
\begin{tabular}{lccc}
\toprule
& \multicolumn{3}{c}{\textbf{Metrics}} \\
\cmidrule{2-4}
\textbf{Algorithm} & Success Rate & Planning Time (\si{\second}) & Simulation Time (\si{\second}) \\ \midrule
\textsc{SPAMP} & 82$\%$ & 2 $\pm$ 4 & 0.8 $\pm$ 0.6 \\ \bottomrule
\end{tabular}
\endgroup
\end{table}


\subsection{In-Depth Analysis of \textsc{SPAMP} in Simulation}\label{subsec:prelimexp}

\begin{table}[]
\caption{Effect of Subgoals and Soft duplicate detection}
\label{tab:ablation1}
\begingroup
\setlength{\tabcolsep}{4pt}
\begin{tabular}{@{}llcccc@{}}
\toprule
\multirow{2}{*}{\textbf{Metrics}} & \multirow{2}{*}{\textbf{Scenario}} & \multicolumn{4}{c}{\textbf{Planning Algorithms}} \\
\cmidrule{3-6}
& & \textsc{Naive} & \textsc{Naive+DD} & \textsc{SubG} & \textsc{SubG+DD}\\ \midrule
\shortstack{Success\\ Rate} & Overall & 90$\%$ & 92$\%$ & 95$\%$ & \textbf{96$\%$} \\ \midrule
\multirow{2}{*}{\shortstack{Planning\\ Time (\si{\second})}} & Easy & 13 $\pm$ 22 & 9 $\pm$ 21 & 7 $\pm$ 14  & \textbf{6 $\pm$ 10}\\
 & Difficult & 433 $\pm$ 388 & 188 $\pm$ 264 & 58 $\pm$ 108 & \textbf{39 $\pm$ 97}\\ \bottomrule
\end{tabular}
\endgroup
\end{table}

To get a better understanding of the quantitative performance of \textsc{SPAMP}, we conducted experiments to highlight the effect of various components. For our first experiment, we highlight the effect of subgoals and soft duplicate detection. We consider four different planning algorithms - \textsc{Naive} is a vanilla MHA* algorithm with only the 3D BFS heuristic to the goal; \textsc{Naive+DD} uses soft duplicate detection on top of the \textsc{Naive} planner, everything else being the same; \textsc{SubG} uses one randomly sampled Phase 1 valid \AMP as a subgoal in MHA*; and \textsc{SubG+DD} uses soft duplicate detection on top of the \textsc{SubG} planner. Problems in this experiment are initialised with 8 movable objects on the tabletop. Table~\ref{tab:ablation1} shows the quantitative benefits of subgoals and soft duplicate detection individually, and that in tandem they can greatly improve performance over the \textsc{Naive} planner, which can be considered a lower-bound on performance for any planning algorithm in this domain.

\begin{table}[]
\caption{Quantitave Performance of \textsc{SPAMP} Variants (Tabletop)}
\label{tab:ablation2}
\begingroup
\setlength{\tabcolsep}{4pt}
\begin{tabular}{@{}llcccc@{}}
\toprule
\multirow{2}{*}{\textbf{Metrics}} & \multirow{2}{*}{\textbf{Scenario}} & \multicolumn{4}{c}{\textbf{Planning Algorithms}} \\
\cmidrule{3-6}
& & \textsc{Naive} & \textsc{Naive+DD} & \textsc{Phase1} & \textbf{\textsc{SPAMP}}\\ \midrule
\shortstack{Success\\ Rate} & Overall & 85$\%$ & 92$\%$ & 97$\%$ & \textbf{99$\%$} \\ \midrule
\multirow{2}{*}{\shortstack{Planning\\ Time (\si{\second})}} & Easy & 16 $\pm$ 22 & 14 $\pm$ 37 & 13 $\pm$ 101  & \textbf{5 $\pm$ 5}\\
 & Difficult & 524 $\pm$ 439 & 379 $\pm$ 461 & 48 $\pm$ 218 & \textbf{11 $\pm$ 16}\\ \midrule
\multirow{2}{*}{\shortstack{Simulation\\ Time (\si{\second})}} & Easy & 7 $\pm$ 13 & 4 $\pm$ 7 & 5 $\pm$ 11  & \textbf{0 $\pm$ 0}\\
 & Difficult & 130 $\pm$ 292 & 63 $\pm$ 136 & 18 $\pm$ 62 & \textbf{1 $\pm$ 7}\\ \bottomrule
\end{tabular}
\endgroup
\end{table}

In a second experiment, we compare the performance of \textsc{SPAMP} against three related variants on the same tabletop workspace experiment from Section~\ref{subsec:sim}. We compare against \textsc{Naive}, \textsc{Naive+DD}, and also a planner (\textsc{Phase1}) which randomly samples $N = 3$ Phase 1 valid \AMP{}s without simulation for use as subgoals. All three of these baselines are allowed to simulate \AMP{}s from within the $\delta$-sphere of a goal configuration from the outset. Table~\ref{tab:ablation2} shows that while simply using Phase 1 valid subgoals has clear benefits over not using them, the necessary reliance on simulating all Phase 1 valid \AMP{}s leads to higher simulation times, and thereby higher planning times, as compared to \textsc{SPAMP}.

\D{All planners in this section utilise Assumption~\ref{assume:amps} from Section~\ref{subsec:assumptions}. If a planning problem is unsolvable due to the assumption being violated, that failure is common to all planners. All other failures for the experiments in this section are due to planners exceeding the timeout $(1800 \si{\second})$.}


\section{Discussion \& Future Work}\label{sec:future}

In this work we present \textsc{SPAMP}, an algorithm for Simulation-based Planning with Adaptive Motion Primitives for the \MAMO domain. We use \AMP{}s as subgoals within a multi-heuristic search framework to solve manipulation planning problems in cluttered scenes. \textsc{SPAMP} improves planning times by up to $40-80\times$ over \textsc{KPIECE}, and up to $2-8\times$ over Selective Simulation, two state-of-the-art baselines for the \MAMO domain. \textsc{SPAMP} also reduces simulation times by up to $20-40\times$ in comparison to these baselines. We show that our assumption of restricting robot-object interactions to terminal \AMP{}s in a plan is not restrictive since we solve $93-99\%$ of all problems. In our future work, we hope to relax our assumption that interactions may only occur during terminal \AMP{}s in order to solve all planning problems in the \MAMO domain.

\bibliographystyle{IEEEtran}
\bibliography{references}

\end{document}